\pdfoutput=1

\documentclass[11pt]{article}

\usepackage[final]{coling}

\usepackage{times}
\usepackage{latexsym}

\usepackage[T1]{fontenc}

\usepackage[utf8]{inputenc}

\usepackage{microtype}

\usepackage{inconsolata}

\usepackage{graphicx}

\usepackage{graphicx}
\usepackage{algorithm}
\usepackage{algorithmic}
\usepackage{booktabs}
\usepackage{makecell}
\usepackage{bm}
\usepackage{hyperref}
\usepackage{amsmath}
\usepackage{amsfonts}
\usepackage{multirow} 
\usepackage{color}
\DeclareMathOperator*{\argmax}{argmax}
\usepackage{newfloat}
\usepackage{listings}

\definecolor{lightgreen}{rgb}{0.55, 0.71, 0.0}
\definecolor{bisque}{rgb}{0.87, 0.72, 0.53}
\definecolor{lightyellow}{rgb}{0.99, 0.76, 0.0}
\definecolor{lightblue}{rgb}{0.36, 0.54, 0.66}
\definecolor{darkgray}{rgb}{0.66, 0.66, 0.66}
\definecolor{salmon}{rgb}{0.98, 0.50, 0.45}

\usepackage[switch]{lineno}

\title{FIPO: Free-form Instruction-oriented Prompt Optimization with Preference Dataset and Modular Fine-tuning Schema}

\author{\makecell{Junru Lu$^{1,2}$, Siyu An$^2$, Min Zhang$^3$, Yulan He$^{1,4,5}$, Di Yin$^2$, Xing Sun$^2$} \\
  $^1$University of Warwick, $^2$Tecent YouTu Lab, $^3$East China Normal University, \\$^4$King's College London, $^5$The Alan Turing Institute\\
    $^1$\texttt{junru.lu@warwick.ac.uk}, $^3$\texttt{mzhang@cs.ecnu.edu.cn}, $^4$\texttt{yulan.he@kcl.ac.uk} \\
  $^2$\texttt{\{siyuan, endymecyyin, winfredsun\}@tencent.com}}

\begin{document}
\maketitle
\begin{abstract}
When carefully optimized by human experts, naive prompts can significantly enhance the task performance of large language models (LLMs). However, such expert-driven prompt optimizations are resource-intensive. To address this, some studies have proposed Automatic Prompt Optimization (APO), which refines naive prompts according to task outputs from in-box testing models, utilizing advanced LLMs (e.g., GPT-4) in an ad-hoc way. 
Although effective, current approaches face challenges in generalization and privacy risks. To overcome these limitations, we have developed the first large-scale Prompt Optimization Preference (POP) dataset, fine-tuned offline local LLM-based optimizers, and conducted fairly evaluations across various downstream models. 
Our method, named Free-from Instruction-oriented Prompt Optimization (FIPO), allows precise optimization of the core task instructions in naive prompts in a model-agnostic manner. 
FIPO uses a modular APO template that dynamically incorporates the naive task instructions, optional instruction responses, and optional ground truth to produce refined prompts. 
The POP dataset is meticulously constructed using advanced LLMs, undergoing rigorous cross-validation by human experts and analytical models. 
By leveraging insights from this dataset, along with Tulu2 models and diverse fine-tuning strategies, we validate the efficacy of the FIPO framework across five public benchmarks and six testing models.\footnote{Our dataset and codes are available at: \url{https://github.com/LuJunru/FIPO_Project}.}
\end{abstract}

\section{Introduction}
Large Language Models (LLMs) have demonstrated impressive capabilities\,\cite{zhao2023survey,yang2023harnessing,openai2023gpt4} across various benchmarks\,\cite{cobbe2021training,suzgun-etal-2023-challenging,bisk2020piqa,huang-etal-2019-cosmos,hendrycks2021measuring}. However, their task performance is highly dependent on the quality of the given task prompt. While LLMs may struggle to produce correct answers when working with naive task prompts, they can excel on the same tasks when guided by  carefully optimized, high-quality prompts crafted by human experts\,\cite{wei2023chainofthought,kojima2022large,yang2023large}.

\begin{figure*}[!t]
  \centering
  \includegraphics[width=1.0\linewidth]{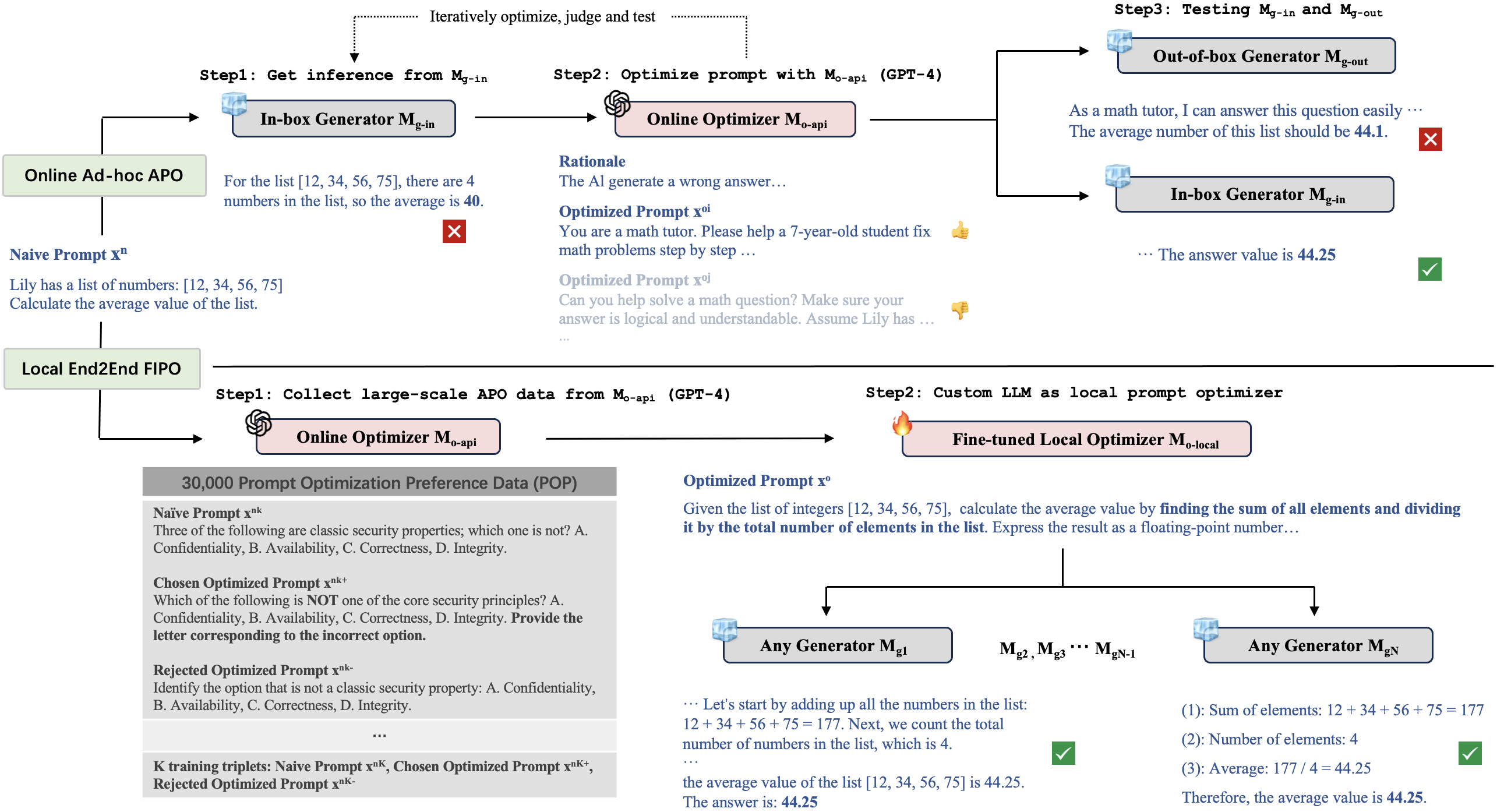}
  \caption{\textbf{Online Ad-hoc APO vs. our Local End-to-End FIPO}: Although both approaches leverage advanced LLMs (e.g., GPT-4), FIPO introduces a a locally trained pipeline that eliminates any dependence on in-box model generators, ensuring a fully self-contained and end-to-end optimization process.}
  \label{fig:example}
\end{figure*}

Obviously, expert-based prompt optimization is costly. 
Therefore, in recent years, Automatic Prompt Optimization (APO) has emerged as a prominent area of research. 
Discrete APO is one of the popular strategies, focusing on identifying optimal combinations of discrete tokens to serve as optimized prompts\,\cite{van-de-kar-etal-2022-dont,yuan2021bartscore,jiang-etal-2020-know,pryzant-etal-2023-automatic}. Particularly, there has been significant interest in LLM-based discrete APO\,\cite{zhou2023large,do2023prompt,wang2023promptagent}, which introduce ad-hoc strategies leveraging leading API-accessible LLMs (e.g., GPT-4\,\cite{openai2023gpt4}). 

These APO approaches typically involve iterative optimization between an in-box testing generator $\bm{M}_{g-in}$ and an advanced optimizer $\bm{M}_{o-api}$. As illustrated in the upper half of Figure\,\ref{fig:example}, the generator $\bm{M}_{g-in}$ first responds to a naive prompt $\bm{x}^n$ such as: ``\emph{Calculate the average value of the list}'', and then the optimizer $\bm{M}_{o-api}$ provides rational feedback and suggests several upgraded prompt candidates $\{\bm{x}^{o}\}$. This iterative process continues until a high-quality optimized prompt $\bm{x}^o$ is generated. This final prompt is tailored to the in-box generator $\bm{M}_{g-in}$, ensuring it produces the desired response, e.g.,  ``\emph{The answer value is 44.25}''. 

Despite their effectiveness, several drawbacks remain: (1) \textbf{Privacy Risk}. The entire online optimization process relies on external LLM services, exposing sensitive information to third-party systems; (2) \textbf{Poor Generalization}. Ad-hoc optimization is highly model-specific, as it depends on immediate testing responses from the in-box generator $\bm{M}_{g-in}$, leading to performance degradation when tested with out-of-box generators $\bm{M}_{g-out}$. For instance, the out-of-box generator $\bm{M}_{g-out}$ might produce an incorrect response like ``\emph{44.1}''.

\begin{figure*}[!t]
  \centering
  \includegraphics[width=1.0\linewidth]{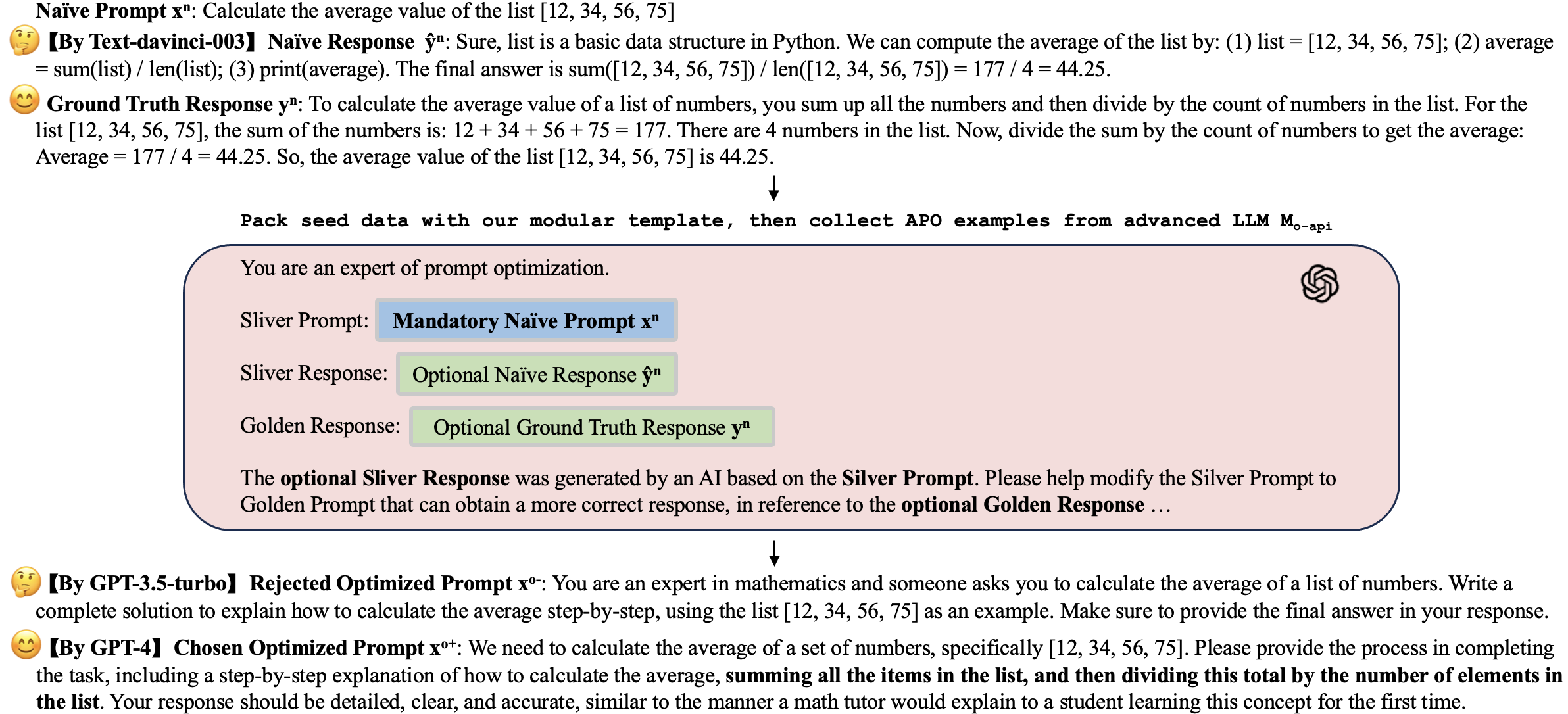}
  \caption{Step 1 and 2 of FIPO: (1) Design a meta-template for universal APO; (2) Collect 30,000 large-scale prompt optimization preference exemples using a suboptimal LLM (GPT-3.5-turbo) and an optimal LLM (GPT-4).}
  \label{fig:data}
\end{figure*}

To address the above limitations, we introduce Free-form Instruction-oriented Prompt Optimization (FIPO). The bottom half of Figure\,\ref{fig:example} illustrates the FIPO framework. Unlike the online ad-hoc APO approach, FIPO directly fine-tunes a general local optimizer $\bm{M}_{local}$, and applies it across any testing generator $\bm{M}_g$. In specific, we first design a meta-template for universal APO (Figure\,\ref{fig:data}), enabling the collection of 30,000 prompt optimization preference examples using an advanced optimizer $\bm{M}_{o-api}$ (\S\,\ref{sec:pop}). We demonstrate the reliability of this dataset through multiple cross-validation methods (Table\,\ref{tab:statistics}). Building on this foundation, we employ mainstream end-to-end fine-tuning strategies to create an effective local optimizer $\bm{M}_{o-local}$ (Figure\,\ref{fig:finetuning}). For example, in Figure\,\ref{fig:example}, the local optimizer $\bm{M}_{local}$ generates an optimized prompt $\bm{x}^o$, providing clear, stepwise guidance: ``\emph{calculate the average value by \textbf{finding the sum of all elements and dividing it by the total number of elements in the list}}''. Such stepwise guidance significantly enhances the answer quality across any testing generator $\bm{M}_g$. Our contributions are as follows:
\begin{itemize}
    \item[(1)] We highlight the drawbacks of previous online ad-hoc APO, and introduce FIPO, a local free-form instruction-oriented prompt optimization. FIFO employs a generic APO template that allows the optimizer to operate independently of pre-built testing generators, a limitation of former methods.
    \item[(2)] Leveraging the APO template, we compile a large-scale Prompt Optimization Preference dataset (POP) with 30,000 examples. We explore several mainstream fine-tuning strategies, including Supervised Fine-tuning (SFT), Direct Preference Optimization (DPO)\,\cite{rafailov2023direct}, and Identity Preference Optimization (IPO)\,\cite{azar2023general}, while also developing an Iterative Preference Learning (IPL) pipeline for enhanced preference optimization.
    \item[(3)] We evaluate the efficacy and adaptability of the fine-tuned FIPO optimizer across five downstream benchmarks and three diverse generators, and obtain superior results compared to existing online ad-hoc APO methods.
\end{itemize}

\section{Methodology}
In this section, we start with task formulation (\S\,\ref{sec:task}), then introduce our meta-template for universal APO (\S\,\ref{sec:modular}), the collected POP data (\S\,\ref{sec:pop}) and the training strategies employed (\S\,\ref{sec:methods}).

\subsection{Task Formulation}
\label{sec:task}
We denote FIPO as end-to-end text generation. In the \textbf{training} phase, a local optimizer model $\bm{M}_{o-local}$ is supervisedly fine-tuned to generate an optimized prompt $\bm{\hat{x}}^{o}$:
\begin{equation}\small
  \label{eqa:fipo_train}
  \bm{\hat{x}}^{o}=\argmax_{\bm{M}_{o-local}} p(\bm{\hat{x}}^{o}|\bm{x}^{n}, [\bm{\hat{y}}^{n}], [\bm{y}^{n}])
\end{equation}
based on the naive prompt $\bm{x}^{n}$, optional naive response $\bm{\hat{y}}^{n}$, and optional ground truth $\bm{y}^{n}$. In addition, pairwise chosen optimized prompt $\bm{x}^{o+}$ and rejected optimized prompt $\bm{x}^{o-}$ are provided as labels in training. The optional naive response $\bm{\hat{y}}^{n}$ is generated for the naive prompt $\bm{x}^{n}$ using one neural generator model $\bm{M}_{g*}$:
\begin{equation}\small
  \label{eqa:fipo_yn}
  \bm{\hat{y}}^{n}=\argmax_{\bm{M}_{g*}} p(\bm{\hat{y}}^{n}|\bm{x}^{n})
\end{equation}
While in the \textbf{testing} phase, our ultimate target is to obtain a more superior optimized testing response $\bm{\hat{y}}^{o}_t$ than the naive testing response $\bm{\hat{y}}^{n}_t$, when applying \underline{any} testing generator $\bm{M}_g$ to the optimized testing prompt $\bm{\hat{x}}^{o}_t$ and the naive testing prompt $\bm{x}^{n}_t$, respectively: 
\begin{equation}\small
  \label{eqa:fipo_test_1}
  \bm{\hat{y}}^{o}_t \succ \bm{\hat{y}}^{n}_t
\end{equation}
\begin{equation}\small
  \label{eqa:fipo_test_2}
  \bm{\hat{y}}^{o}_t=\argmax_{\bm{M_g}} p(\bm{\hat{y}}^{o}_t|\bm{\hat{x}}^{o}_t)\mbox{, }\bm{\hat{y}}^{n}_t=\argmax_{\bm{M_g}} p(\bm{\hat{y}}^{n}_t|\bm{\hat{x}}^{n}_t)
\end{equation}
where $\bm{M}_{g}$ could be either same as or different from $\bm{M}_{g*}$. And specifically, $\bm{\hat{x}}^{o}_t$ is enhanced from $\bm{x}^{n}_t$ by the fine-tuned optimizer $\bm{M}_{o-local}$:
\begin{equation}\small
  \label{eqa:fipo_test_3}
  \bm{\hat{x}^{o}_t}=\argmax_{\bm{M_o}} p(\bm{\hat{x}}^{o}_t|\bm{x}^{n}_t)
\end{equation} 

In contrast, former ad-hoc APO has no training phase but only the iterative online testing pipeline with mandatory \underline{in-box} testing response $\bm{\hat{y}}^{o_i}_t$:
\begin{equation}\small
  \label{eqa:apo_test_1}
  \bm{\hat{x}}^{o_{i+1}}_t=\argmax_{\bm{M}_{o-api}} p(\bm{\hat{x}}^{o_{i+1}}_t|\bm{x}^{o_i}_t,\bm{\hat{y}}^{o_i}_t)\mbox{, }\bm{x}^{o_1}_t=\bm{x}^{n}_t
\end{equation} 
\begin{equation}\small
  \label{eqa:apo_test_2}
  \bm{\hat{y}}^{o_i}_t=\argmax_{\bm{M}_{g-in}} p(\bm{\hat{y}}^{o_i}_t|\bm{x}^{o_i}_t)\mbox{, }\bm{\hat{y}}^{o_1}_t=\bm{\hat{y}}^{n}_t
\end{equation} 
where $\bm{M}_{g-in}$ is a prior in-box generator.

\subsection{Modular Template}
\label{sec:modular}
As aforementioned in section\,\ref{sec:task}, we first design a modular template that ensures flexibility in content management. 
Figure\,\ref{fig:data} illustrates our template, shown in the middle, taking {\color{lightblue}{mandatory naive task instruction $\bm{x}^{n}$}}, {\color{lightgreen}{optional naive response $\bm{\hat{y}}^{n}$}} and {\color{lightgreen}{optional ground truth response $\bm{y}^{n}$}} as inputs. Additional description is then appended for clarity, in which we directly claim the optionality of $\bm{\hat{y}}^{n}$ and $\bm{y}^{n}$: ``\emph{The \textbf{optional} sliver response ··· based on the sliver prompt ··· \textbf{optional} golden response ···}''. 

We use this modular template for all sections in FIPO, including the dataset collection, fine-tuning the local optimizer $\bm{M}_{o-local}$, and testing various downstream generators $\{\bm{M}_{g}\}$. The key difference between these sections is we accordingly adjust the optional responses, thus addressing potential exposure bias between the training and the inference phases (Eq.\,\ref{eqa:fipo_train} vs. Eq.\,\ref{eqa:fipo_test_3}): (1) \textbf{Keep both in data collection}. We introduce the collection of our POP data in section\,\ref{sec:pop}; (2) \textbf{Diversely keep partial responses in training}. We present this strategy in section\,\ref{sec:methods}; (3) \textbf{Remove all responses in testing}. We report testing results in section\,\ref{sec:exps}.

\begin{figure*}[!t]
  \centering
  \includegraphics[width=1.0\linewidth]{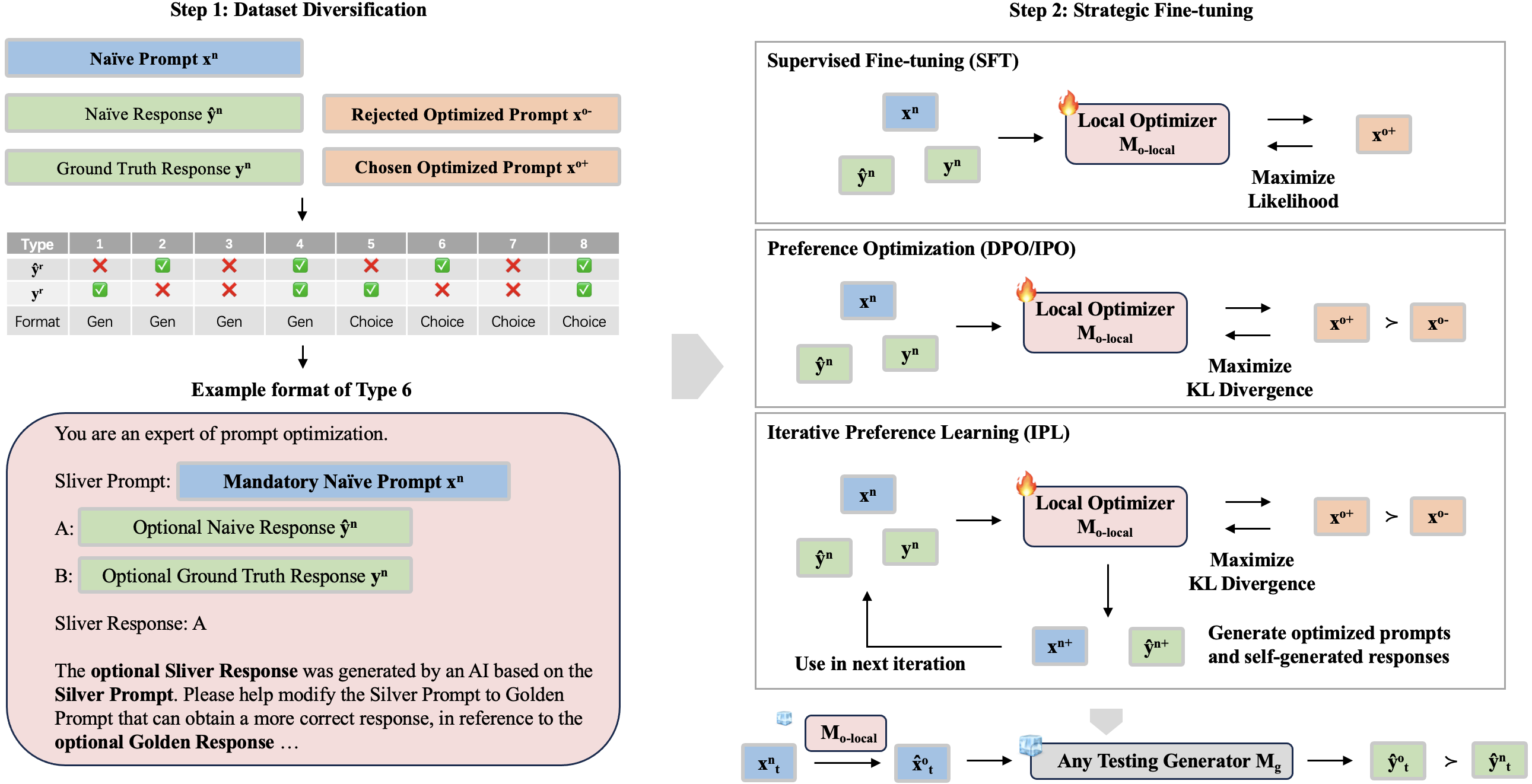}
  \caption{Step 3 of FIPO: transitional dataset diversification and several mainstream fine-tuning strategies.}
  \label{fig:finetuning}
\end{figure*}

\subsection{Prompt Optimization Preference Data}
\label{sec:pop}
We decide to distill the refined prompt optimization capabilities from prominent yet proprietary LLMs, instead of directly integrating them in an ad-hoc way. Thus, we collect the Prompt Optimization Preference (POP) data. Shown in Figure\,\ref{fig:data}, to ensure the most directional optimization, we send naive prompt $\bm{x}^{n}$, naive response $\bm{\hat{y}}^{n}$, and ground truth response $\bm{y}^{n}$ to one suboptiomal LLM GPT-3.5-turbo and one optimal LLM GPT-4\footnote{\url{https://platform.openai.com/docs/models}}, to collect contrastive POP data ($\bm{x}^{o+}$, $\bm{x}^{o-}$). 
The naive prompt $\bm{x}^{n}$ is sampled from the Alpaca dataset, which contains 52K diverse instructions and corresponding responses $\bm{\hat{y}}^{n}$ generated by the Text-davinci-003 model\,\cite{taori2023stanford}. We also collect another GPT4-generated response for the Alpaca dataset from public literature\,\cite{ghosal2023flacuna}. There are no official ground truth responses (e.g., from human experts) for the Alpaca data, we therefore consider GPT-4 responses as the ground truth response $\bm{y}^{n}$, given its demonstrated analytical capabilities comparable to humans\,\cite{pan2023rewards}. 
As shown at the bottom of Figure\,\ref{fig:data}, GPT-4 offers a more pedagogical step-by-step optimized prompt compared to GPT-3.5-turbo. We report complete collection template in Table\,\ref{tab:collection}.

We finally narrow down our dataset to 30k samples and report the quality post-checking in Table\,\ref{tab:statistics}. We adopt cross-validation using three different methods: critiques from an external alignment model UltraRM\,\cite{cui2023ultrafeedback}, self-judgement from the GPT-4, and manual checking by human experts. The ``\emph{Response}'' and ``\emph{Prompt}'' columns refer to the proportions that GPT-4-generated response and GPT-4-optimized prompt is better than the others, respectively. The average win rates for both categories exceed 85\%, ensuring the quality.

\begin{table}[t]
\resizebox{\columnwidth}{!}{%
  \begin{tabular}{l|c|c|c}
    \toprule
    \quad & \multicolumn{2}{c|}{GPT-4 Win Rate (\%)} & \quad \\
    \quad & \textbf{Response} & \textbf{Prompt} & \textbf{Scale} \\
    \midrule
    \textbf{UltraRM 13B\,\cite{cui2023ultrafeedback}} & 91.49 & 82.13 & 30k \\
    \midrule
    \textbf{GPT4 Self-check} & 80.56 & 92.29 & 3k \\
    \midrule
    \textbf{Human Expert} & 88.29 & 95.21 & 1k \\
    \midrule
    \textbf{Average} & 86.78 & 89.88 & N/A \\
    \bottomrule
  \end{tabular}}
    \caption{Quality cross-validation on our dataset.}
  \label{tab:statistics}
\end{table}

\subsection{Fine-tuning Strategies}
\label{sec:methods}
We introduce our fine-tuning strategies in Figure\,\ref{fig:finetuning}, consisting of an initial step of 
transitional dataset diversification followed by strategic fine-tuning.

\noindent \textbf{Dataset Diversification.} In the left hand of Figure\,\ref{fig:finetuning}, we evenly split the 30k samples as eight types depending on the existence of naive response $\bm{\hat{y}}^{n}$ and ground truth response $\bm{y}^{n}$, as well as a format condition ``\emph{generation}'' or ``\emph{multi-choice}''. Directionally fine-tuning optimizer has to relay on pre-generated responses, while no any response will be exposed during inference. Hence, the dataset diversification is necessary to help reduce the exposure gap between training and testing, and generalize the original ``\emph{generation}'' instruction format to another common ``\emph{multi-choice}'' instruction format. The left bottom corner in Figure\,\ref{fig:finetuning} takes an example of Type 6. The responses $\bm{\hat{y}}^{n}$ and $\bm{y}^{n}$ are modified as candidates adhere to the naive prompt $\bm{x}^{n}$. We then set ``\emph{A}'' and ``\emph{B}'' as new naive response $\bm{\hat{y}}^{n}$ and ground truth response $\bm{y}^{n}$\footnote{More details of data diversification are in Appendix\,\ref{sec:app_data}.}.

\noindent \textbf{Strategic Fine-tuning.} The right side of Figure\,\ref{fig:finetuning} introduces several end-to-end fine-tuning strategies that we explored in this work. The right top is the most well-known Supervised Fine-tuning (SFT), which only takes the optimal optimized prompt $\bm{x}^{o+}$ as the supervision signal:
\begin{equation}\small
  \label{eqa:sft}
  L_{SFT}(M_o)=-\mathbb{E}_{(x^n,\hat{y}^n,y^n,x^{o+})\sim D}[\hat{x}^{o} - x^{o+}]^2
\end{equation}
where $D$ stands for the training set.

On the other hand, the right middle shows a contrastive fine-tuning methodology: Preference Optimization, such as Direct Preference Optimization (DPO)\,\cite{rafailov2023direct} and Identity Preference Optimization (IPO)\,\cite{azar2023general}. Preference Optimization takes pairwise rejected label $\bm{x}^{o-}$ and chosen label $\bm{x}^{o+}$ as supervision. One of the core differences from Preference Optimization and SFT is the former one not only encourage the generation of optimal preference, but also dampen the generation of suboptimal preference:
\begin{equation}\small
  L_{DPO}(M_o)=-\mathbb{E}_{(x^n,\hat{y}^n,y^n,x^{o+},x^{o-})\sim D}[\log\sigma(\beta\cdot\Delta)]
\end{equation} 
\begin{equation}\small
  \label{eqa:ipo}
  L_{IPO}(M_o)=-\mathbb{E}_{(x^n,\hat{y}^n,y^n,x^{o+},x^{o-})\sim D}[\Delta-\frac{1}{2\beta}]^2
\end{equation} 
\begin{equation}\small
  \label{eqa:po}
  \Delta=\log\frac{M_o(x^{o+}|x^r,\hat{y}^r,y^r)}{M_{ref}(x^{o+}|x^r,\hat{y}^r,y^r)} - \log\frac{M_o(x^{o-}|x^r,\hat{y}^r,y^r)}{M_{ref}(x^{o-}|x^r,\hat{y}^r,y^r)}
\end{equation}
where $\bm{\beta}$ is a hyperparameter factor. $\bm{M_{ref}}$ refers to the reference model, which is a frozen copy of initial weights of $\bm{M_o}$. The equations indicate that IPO is a regularized version of DPO as it limits the optimization range with squares.

Additionally, inspired by self-updating alignment\,\cite{lee2023rlaif,bai2022constitutional,yuan2024self}, we develop a Iterative Preference Learning (IPL) strategy for self-rewarding prompt optimization. After each iteration of prompt optimization, we ask the optimizer itself to determine if it successfully generate a superior prompt $\bm{x}^{n+}$ with a better response $\bm{\hat{y}}^{n+}$, and if so, to automatically replace the previous inferior prompt $\bm{x}^{n}$ and response $\bm{\hat{y}}^{n}$, leading to more rigorous training in next iteration:
\begin{equation}\small
  \label{eqa:ipl}
  L_{IPL}(M_o)=-\mathbb{E}_{(x^{n+},\hat{y}^{n+},y^n,x^{o+},x^{o-})\sim D}G(\Delta)
\end{equation}
\begin{equation}\small
  \label{eqa:ipl_op}
  x^{n+}=M_o(x^n)\mbox{, }\hat{y}^{n+}=M_o(x^{n+})
\end{equation}
\begin{equation}\small
  \label{eqa:ipl_judge}
  x^{n+} = \left\{
  \begin{array}{l}
      x^{n+}\mbox{, }M_o(x^{n+}, y^n) \succ M_o(x^n, y^n) \\
      x^{n}\mbox{, otherwise}
  \end{array}
  \right.
\end{equation}
where $G(*)$ denotes as either IPO or DPO loss\footnote{Algorithmic details of IPL lie in Appendix\,\ref{sec:app_data} and \,\ref{app:ipl}.}.

\section{Experiments}
The ultimate target of FIPO lies in the general performance enhancement with downstream generators $\bm{M}_g$ (Eq.\,\ref{eqa:fipo_test_1}). Herein, in the evaluation, we first use fine-tuned optimizer $\bm{M}_o$ to produce optimized testing prompts $\bm{\hat{x}}_{t}^o$, then obtain optimized testing response $\bm{\hat{y}}_t^o$ and naive testing response $\bm{\hat{y}}^{n}_t$ for answer quality checking, as shown in the bottom right corner of Figure\,\ref{fig:finetuning}. We begin with our experimental settings (\S\,\ref{sec:setup}), efficacy presentation and comparisons against online ad-hoc APO methods (\S\,\ref{sec:sota}), then follow with analysis of different fine-tuning strategies (\S\,\ref{sec:fine_analyasis}), and case analysis (\S\,\ref{sec:case})\footnote{Hyperparameters and training cost are in Appendix\,\ref{sec:app_cost}.}.

\subsection{Experimental Settings}
\label{sec:setup}

\subsubsection{Baselines} 
We compare FIPO with two SOTA APO methods: APE\,\cite{zhou2023large} and PromptAgent\,\cite{wang2023promptagent}. APE, stands for Automatic Prompt Engineer, is a template-based strategy that ask one LLM to generate a pool of candidate prompts based on the templates, then select one according to the evaluation scores. PromptAgent eliminates templates and replaces with Monte Carlo Tree Search\,\cite{abramson2014expected} for using a evaluator model to guide the generator. Both APE and PromptAgent are training-free, aiming to model-oriented APO in an ad-hoc manner, while we realize completely offline training. Following former works, we use GPT-3.5-turbo as the in-box generator and GPT-4 as the optimizer in both baselines.

We use Tulu2 models as our bases, which is a fine-tuned version of Llama2\,\cite{touvron2023llama} trained on a mix of publicly available datasets\,\cite{ivison2023camels}. We fine-tuning local optimizer with Tulu2-13B and Tulu2-70B.

\begin{table*}[!t]
\resizebox{\textwidth}{!}{%
  \begin{tabular}{c|c|cc|ccc|c}
    \toprule
    \quad & \quad & \multicolumn{2}{c|}{\textbf{Generation}} & \multicolumn{3}{c|}{\textbf{Multi-choice}} & \quad\\
    \toprule
    \textbf{Generator} & \textbf{Prompt Source} & \textbf{GSM8K (3)} & \textbf{BBH (3)} & \textbf{PiQA (3)} & \textbf{CosmosQA (5)} & \textbf{MMLU (5)} & \textbf{Weighted Avg.}\\
    \midrule
    Llama2-7B & Naive Prompt & 8.89 & 31.21 & 62.78 & 43.09 & 46.58 & 41.73 \\
    \,\cite{touvron2023llama} & FIPO Optimizer & \textbf{11.70} & \textbf{33.50} & \textbf{69.37} & \textbf{52.11} & \textbf{54.56} & \textbf{48.10} \\
    \midrule
    Tulu2-13B & Naive Prompt & 39.06 & 36.49 & 76.62 & 55.13 & 57.43 & 52.53 \\
    \,\cite{ivison2023camels} & FIPO Optimizer & \textbf{40.17} & \textbf{40.26} & \textbf{78.58} & \textbf{57.68} & \textbf{59.10} & \textbf{54.79} \\
    \midrule
    Baichuan2-13B & Naive Prompt & 46.81 & 37.95 & 68.56 & 51.88 & 57.46 & 52.36 \\
    \,\cite{yang2023baichuan} & FIPO Optimizer & \textbf{48.12} & \textbf{39.95} & \textbf{74.77} & \textbf{56.88} & \textbf{58.32} & \textbf{54.35} \\
    \bottomrule
  \end{tabular}}
    \caption{Evaluation results of various downstream generator LLMs, using the best local optimizer from Table\,\ref{tab:short_finetune}.}
  \label{tab:generators}
\end{table*}

\begin{figure*}[!t]
  \centering
  \includegraphics[width=1.0\linewidth]{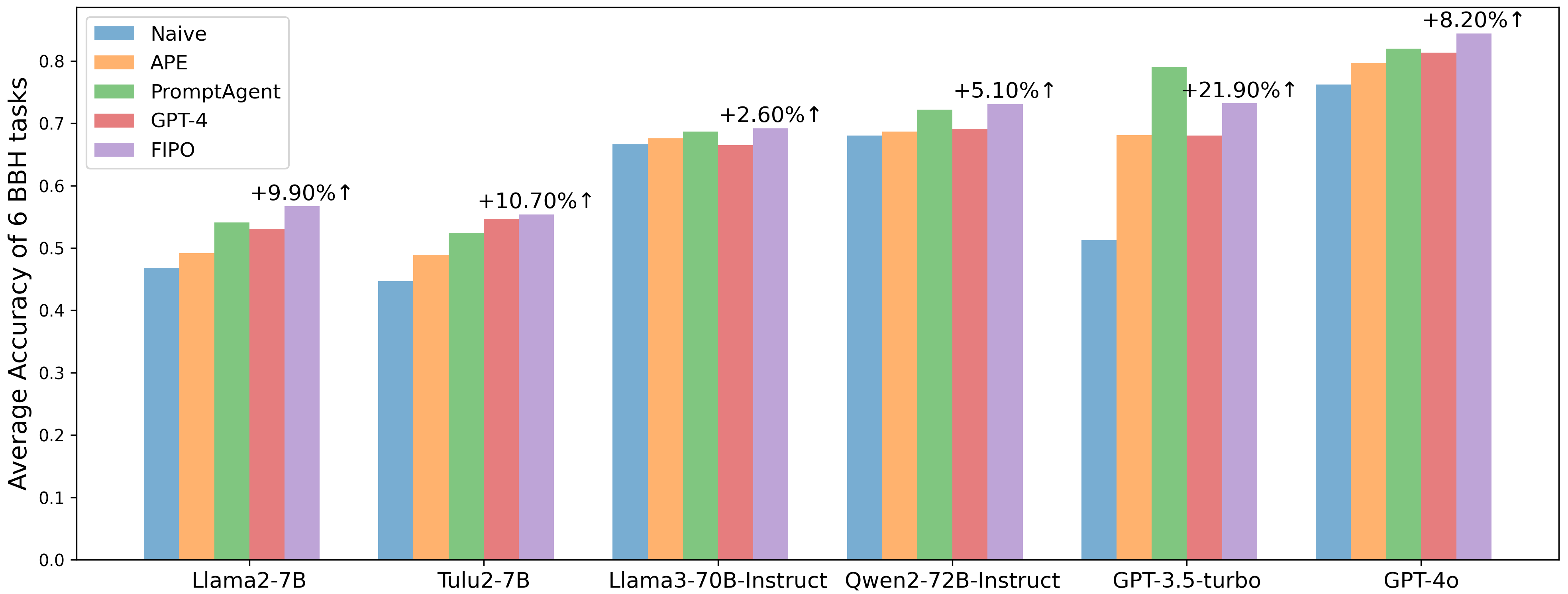}
  \caption{The FIPO-optimized prompts help various downstream testing LLMs (\textbf{X-axis}) gain more promising improvements, compared with other prompt optimization approaches (shown by the \textbf{bars}). We specifically annotate the improvements of FIPO against naive prompts from the original dataset (\textbf{↑}). More details: Appendix\,\ref{app:cost} and \,\ref{app:bbh}.}
  \label{fig:bbh_results}
\end{figure*}

\subsubsection{Evaluation Benchmarks} 
We include five benchmarks across two most common formats: (1) GSM8k\,\cite{cobbe2021training}, a generative dataset contains 1.3k primary level math questions; (2) BigBenchHard (BBH)\,\cite{suzgun-etal-2023-challenging}, which involves 23 challenging reasoning tasks. BBH has 6.4k testing samples, and asks for generative answering; (3) PiQA\,\cite{bisk2020piqa}, in which 1.8k common physical knowledge questions are proposed, alongside with multiple candidate choices; (4) CosmosQA\,\cite{huang-etal-2019-cosmos}. There are around 3k commonsense-based multi-choice questions in CosmosQA, equipped with four candidate options; (5) MMLU\,\cite{hendrycks2021measuring}, which is one of the largest multi-choice benchmarks. MMLU covers 14k questions. For our FIPO experiments, we report results on all five benchmarks. While differently, since both APE and PromptAgent only provide evaluations on 6 tasks of BBH, we report the comparison results aligned with their settings. As for the result metrics, either ``\emph{generation}'' or ``\emph{multi-choice}'' benchmarks takes few-shot format with strict answering templates (e.g., ``\emph{The answer is X}''). Herein, we are able to report the accuracy score for all benchmarks.

\subsection{Experimental Results}
\label{sec:exps}

\subsubsection{Efficacy of FIPO} 
\label{sec:sota}
\quad \textbf{General improvements of FIPO}. It can be concluded that FIPO-optimized prompts have general gains on different downstream generators across five public benchmarks, shown in Table\,\ref{tab:generators}. The optimized prompts help Llama2-7B, Tulu2-13B and Baichuan2-13B models gain 6.37\%, 2.26\% and 1.99\% performance growth on average. 

\begin{table*}[!t]
\resizebox{\textwidth}{!}{%
  \begin{tabular}{c|cc|ccc|c}
    \toprule
    \quad & \multicolumn{2}{c|}{\textbf{Generation}} & \multicolumn{3}{c|}{\textbf{Multi-choice}} & \quad\\
    \toprule
    \textbf{FIPO Prompt Optimizer} & \textbf{GSM8K (3)} & \textbf{BBH (3)} & \textbf{PiQA (3)} & \textbf{CosmosQA (5)} & \textbf{MMLU (5)} & \textbf{Weighted Avg.}\\
    \midrule
    Naive Prompt & 24.77 & 36.21 & 73.35 & 51.17 & 51.22 & 47.79 \\
    \midrule
    Best 13B Optimizer & 18.42 & 33.55 & 72.03 & 49.14 & 48.20 & 44.93 \\
    \midrule
    SFT-70B & 21.43 & 32.92 & 74.39 & 49.97 & 51.55 & 46.96 \\
    DPO-70B & \textbf{27.74} & 35.56 & 74.17 & 54.93 & 52.73 & 49.07 \\
    IPO-70B & 25.00 & 39.21 & 76.84 & 56.01 & 54.29 & 50.94 \\
    IPL-DPO-70B & 25.13 & 35.25 & 74.95 & 50.46 & 52.12 & 48.10 \\
    IPL-IPO-70B & 26.67 & 39.60 & \textbf{77.11} & \textbf{56.71} & \textbf{56.02} & \textbf{52.13} \\
    \midrule
    IPO-70B-gen & 22.72 & \textbf{41.91} & 74.53 & 53.60 & 52.74 & 50.23\\
    IPO-70B-partial & 23.08 & 40.03 & 76.22 & 54.29 & 51.99 & 49.59\\
    \bottomrule
  \end{tabular}}
    \caption{Evaluation results of FIPO optimizer fine-tuned with different strategies, tested with same Tulu2-7B model. The number attached with the benchmark is the number of in-context examples (e.g., BBH (3) means 3-shot testing on BBH). FIPO only optimizes the last task instruction, leaving in-context examples remained.}
  \label{tab:short_finetune}
\end{table*}

\textbf{Comparable optimization capability against online ad-hoc APO, even better}. We would like to compare the local FIPO optimizer with previous methods and direct prompt optimization using GPT-4. Figure\,\ref{fig:bbh_results} reports experimental results on six BBH tasks, following the experimental settings in PromptAgent work\,\cite{wang2023promptagent}. Our FIPO method takes the lead in all downstream tests, except for the in-box tester GPT-3.5-turbo, which is ad-hocly included during the iterative prompt optimization in APE and PromptAgent. In specific, the final average improvements on two open-source 70B models are around 3\% to 5\%, compared with more than 10\% gains on two open-source 7B models. We can notice that as the tested open-soruce model grows larger and stronger, the effectiveness of all prompt optimization methods significantly decreases, which may be due to the firmness of the larger model's inherent knowledge. As for proprietary GPT-3.5 and GPT-4, we found that prompt optimization seems to be more beneficial to them. Prompts optimized by our FIPO can help GPT3.5 improve the final average effect up to 22\%, and maintain an effect of about 8\% on GPT4o.

\subsubsection{Fine-tuning Analysis} 
\label{sec:fine_analyasis}
We fine-tune Tulu2-13B and -70B models as our local optimizer through different strategies as mentioned above. We use downstream performance of Tulu2-7B for analyzing the effectiveness of different fine-tuning strategies. Our findings are here:

\textbf{Small optimizer fails}. ``\emph{Small}'' Tulu2-13B are not up to the difficult prompt optimization task (the 4th line of Table\,\ref{tab:short_finetune}). The average testing scores of using optimized prompts even worse than using the naive prompts written by human.

\textbf{SFT vs. DPO/IPO vs. IPL}. When simply provide a best optimized prompt as the supervision label, indicated by SFT-70B results, the end-to-end prompt optimization is still a hard task. While when contrastive preference supervisions are provided, there are promising improvements obtains, ranging from marginal 0.31\% to significant 4.34\%. In terms of different preference fine-tuning methods, IPO beats DPO in either solely fine-tuning, or joint integration in our proposed IPL pipeline, which may due to its regularized design in Eq.\,\ref{eqa:ipo}. We analyze more fine-tuning details and the self-rewarding benefits of IPL in Appendix\,\ref{sec:app_finetune}.

\textbf{Dataset diversification is necessary}. In the bottom of Table\,\ref{tab:short_finetune}, we present ablation studies of preprocessing dataset diversification, mentioned in section\,\ref{sec:methods}. In specific, IPO-70B-gen stands for not diversifying half of the training set into multi-choice format, which is introduced as type 5,6,7 and 8 in Figure\,\ref{fig:finetuning}. As for IPO-70B-partial, we only use type 3,4,7 and 8 in Figure\,\ref{fig:finetuning} as pairwise diversification templates. The ablated optimizer weakens all benchmarks, except BBH, which is due to its unique symbolic reasoning pattern (Appendix\,\ref{app:ablation}).

\begin{table*}[!t]\small
\resizebox{\textwidth}{!}{%
  \begin{tabular}{p{\textwidth}}
    \toprule
    \textbf{Naive Testing Prompt I $\bm{x}^{n}_{t_1}$ from BBH}: \\
    Jane was born on the last day of February in 2000. Today is her 16-year-old birthday. What is the date a month ago in MM/DD/YYYY?\\
    \textbf{Optimized Testing Prompt I $\bm{x}^{o}_{t_1}$ by FIPO} \\
    As today marks Jane's 16th birthday, determine the date from exactly one month prior. Jane was born on the last day of February in the year 2000, {\color{lightblue}{\textbf{a leap year}}}. Ensure your response is in the format MM/DD/YYYY.\\
    \midrule
    \textbf{Naive Testing Prompt II $\bm{x}^{n}_{t_2}$ from MMLU}: \\
    Three of the following are classic security properties; which one is not? A. Confidentiality, B. Availability, C. Correctness, D. Integrity.\\
    \textbf{Optimized Testing Prompt II $\bm{x}^{o}_{t_2}$ by FIPO} \\
    Which of the following is {\color{lightblue}{\textbf{NOT}}} one of the core security principles? A. Confidentiality, B. Availability, C. Correctness, D. Integrity. {\color{lightblue}{\textbf{Provide the letter corresponding to the incorrect option.}}}\\
    \midrule
    \textbf{Naive Testing Prompt III $\bm{x}^{n}_{t_3}$ from CosmosQA}: \\
    A team of engineers constructing signal lights for airplanes that they can use to guide them to runways are attempting to determine the brightness needed for the pilot to be able to detect the tower at 1 mile away. They set the light to a test brightness and establish communication with an inbound pilot. When the pilot is 1 mile away from the tower, he says he cannot see the light. In terms of Signal Detection Theory, what would this be called?\\
    \textbf{Optimized Testing Prompt III $\bm{x}^{o}_{t_3}$ by FIPO} \\
    According to Signal Detection Theory, {\color{lightblue}{\textbf{when a pilot cannot detect a signal at a set brightness level, it is called a 'miss'. In contrast, it is called a 'hit'.}}} A team of engineers is testing the brightness of signal lights for airplanes to guide them to runways. They establish communication with an inbound pilot and set the light to a test brightness. When the pilot is one mile away from the tower, he reports that he cannot see the light. What term from Signal Detection Theory describes this situation?\\
    \midrule
    \textbf{Naive Testing Prompt IV $\bm{x}^{n}_{t_4}$ from GSM8K}: \\
    Lily has a list of numbers: [12, 34, 56, 75]. Calculate the average value of the list.\\
    \textbf{Optimized Testing Prompt IV $\bm{x}^{o}_{t_4}$ by FIPO} \\
    To find the average of the given list of numbers, first, you need to {\color{lightblue}{\textbf{add all the numbers in the list}}}, which are: 12, 34, 56, and 75. Add them up, and you will get \underline{\textbf{a sum of 177}}. Then, {\color{lightblue}{\textbf{divide the sum by the total number of items in the list}}}, \underline{\textbf{which is 4}}. So, 177 divided by 4 \underline{\textbf{equals 44.25}}. Therefore, the average of the list [12, 34, 56, 75] is 44.25.\\
    \midrule
    \bottomrule
  \end{tabular}}
    \caption{Examples from testing benchmarks.}
  \label{tab:case}
\end{table*}

\subsubsection{Case Analysis.} 
\label{sec:case}
In Table\,\ref{tab:case}, we present several examples from the downstream testing benchmarks, discussing the efficacy and shortcomings of FIPO. Particularly, we smear ({\color{lightblue}{key optimized content with blue}}), and overwhelmed (\underline{\textbf{cheating notes with underlines}}). The 1st optimized prompt of BBH case explicitly mentions that ''\emph{2000 is a leap year}'', which is a key detail for calculating dates in February. The 2nd optimized prompt of MMLU question capitalizes ``\emph{NOT}'' to draw attention to the negative aspect, ensuring the model focuses on identifying the incorrect option. It also explicitly instructs the model to provide the letter of the incorrect option, reducing ambiguity. And the 3rd optimized prompt of CosmosQA case provides definitions for ``\emph{miss}'' and ``\emph{hit}'' according to Signal Detection Theory, making it easier for the model to understand the correct term. While in the last GSM8K case, FIPO breaks down the calculation into clear, step-by-step instructions, ensuring the model understands the process of finding the average. However, it provides overwhelmed cheating notes of the final answer\footnote{This issue mostly occurs in GSM8K and BBH's math calculation questions, happening <10\%, in Appendix\,\ref{sec:app_cheat}.}.

\section{Related Work}
\noindent \textbf{Automatic Prompt Optimization} (APO) is an simple yet effective technique for grabbing potentials of LLMs in various downstream scenarios. Most APO methodologies can be categorized into two types: discrete APO and continuous APO\,\cite{liu2023pre}. Discrete APO searches optimized prompts with optimal combinations of discrete tokens\,\cite{wallace-etal-2019-universal,shin-etal-2020-autoprompt,ben2022pada,davison-etal-2019-commonsense,deng-etal-2022-rlprompt,zhang2023tempera,xu-etal-2022-gps}. For instance, \,\cite{van-de-kar-etal-2022-dont} employed text mining for searching candidate prompts from knowledge triplets. While \,\cite{yuan2021bartscore}, \,\cite{haviv-etal-2021-bertese} and \,\cite{gao-etal-2021-making} utilized Bart\,\cite{lewis2019bart}, Bert\,\cite{devlin2018bert} and T5\,\cite{raffel2019exploring} for optimizing prompts in an paraphrasing manner, respectively.

In contrast, continuous APO proposes to search better prompts in continuous embedding space rather than limited to human-understandable discrete tokens\,\cite{tsimpoukelli2021multimodal,zhong-etal-2021-factual,qin-eisner-2021-learning,hambardzumyan-etal-2021-warp,wang2023multitask}. Prefix Tuning\,\cite{li-liang-2021-prefix} and Prompt Tuning\,\cite{lester-etal-2021-power} are two well-known continuous APO methods that insert prefix vectors in front of the task sequence, and then update their corresponding parameters. There are also hybrid works to insert continuous embedding into discrete templates\,\cite{liu2021gpt,han2021ptr}.

\noindent \textbf{Preference Optimization for LLMs} allows LLMs to align with human minds in a more nuanced way\,\cite{pan2023automatically}, compared to SFT. Proximal Policy Optimization (PPO) is one of the well-known preference optimization approach to first train a reward model with pairwise human preference data, then align LLMs with the reward model via reinforcement learning\,\cite{ouyang2022training,bai2022training}. Despite its efficacy, PPO is often blamed for its training instability and expensive costs. To this end, Direct Preference Optimization (DPO) have been proposed, aiming to align LLMs through implicit modeling, thereby eliminating the flaws associated with the explicit use of reward models\,\cite{rafailov2023direct}. Following works of DPO are presented\,\cite{azar2023general,zhao2023slic,ethayarajh2024kto,lu2024eliminating}.

\section{Conclusion}
We introduce FIPO, the Free-form Instruction-oriented Prompt Optimization. The modular FIPO template proposes to address APO as end-to-end text generation, flexibly taking naive prompt, naive response and ground truth as inputs, for obtaining a new optimized prompt. We hereby collect a large-scale prompt optimization preference dataset, employ with multiple fine-tuning strategies, and then validate the efficacy across objective benchmarks with various downstream generators.

\section*{Limitations}
While FIPO demonstrates significant potential in optimizing prompts for various downstream tasks, there are several limitations to consider:

\textbf{(1) Overwhelmed Cheating Notes}. As shown in the case analysis, FIPO sometimes provides overly detailed instructions that can be considered as ``cheating notes''. This issue is particularly prevalent in tasks involving mathematical calculations. While this enhances performance, it may not align with the intended use of prompt optimization. \textbf{(2) Evaluation Metrics}. The current evaluation primarily focuses on accuracy metrics. While accuracy is important, other aspects such as the interpretability, fairness, and ethical implications of the optimized prompts should also be considered in future work. \textbf{(3) Optimization of In-context Examplars}. FIPO does not include optimization of the in-context examples, but rather focuses on the optimization for the task instructions only.

\section*{Acknowledgements}
This work was supported in part by the UK Engineering and Physical Sciences Research Council (EPSRC) through a Turing AI Fellowship (grant no. EP/V020579/1,
EP/V020579/2), and Innovate UK through the Accelerating Trustworthy AI programme (grant no. 10093055).

\bibliography{coling_latex}

\appendix

\begin{figure*}[!t]
  \centering
  \includegraphics[width=1.0\linewidth]{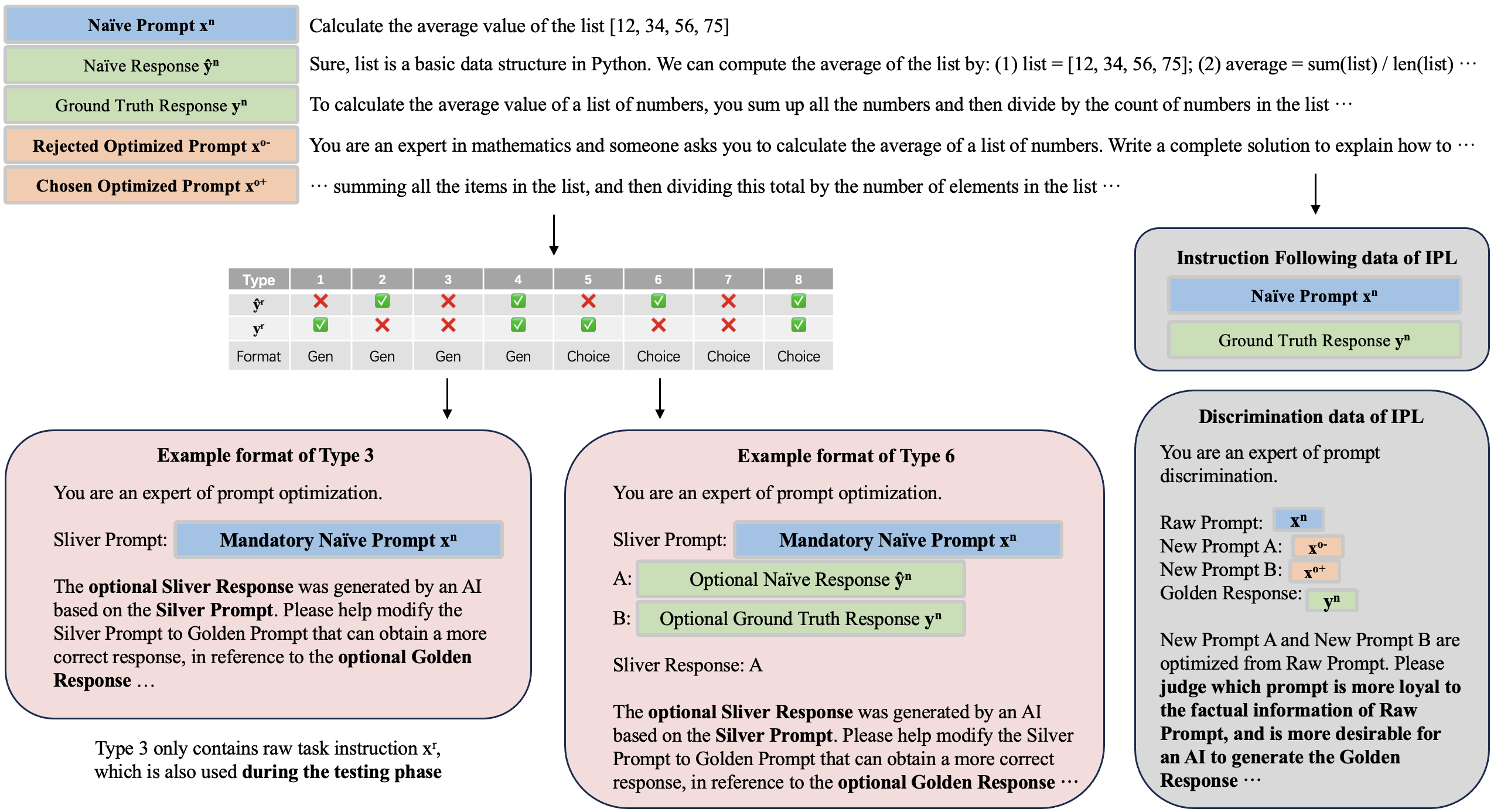}
  \caption{An overview of our dataset diversification step. It is recommended to view the details with colors.}
  \label{fig:diversification}
\end{figure*}

\section{Dataset Diversification}
\label{sec:app_data}
We present our complete data diversification plans in Figure\,\ref{fig:diversification}. We first collect response and prompt preference data as shown in the top, according to pipeline introduced in section\,\ref{sec:methods}. Afterwards, to better minimize the exposure gaps, we carry on the diversification combinations as shown within the table in the left middle of Figure\,\ref{fig:diversification}. The diversification optionally selects naive response $\bm{\hat{y}^n}$, ground truth response $\bm{y}^n$ and ``generation'' or ``multi-choice'' format. We thereby have eight various data format ($2 \times 2 \times 2$), and we demonstrate with two types in the bottom left and middle parts. The left corner one is an example format of type 3, in which only mandatory naive prompt $\bm{x}^n$ is included. It worth to mention that this format type is also used during the testing phase as no responses are provided at that stage. Another type 6 located in the middle bottom transforms the initial generation format as a multi-choice one, and add a new binary naive response $\bm{\hat{y}^n}$. Nevertheless, either type 3 or type 6 has same task description attached.

The right side of Figure\,\ref{fig:diversification} illustrates specific incremented data for IPL fine-tuning. Equations\,\ref{eqa:po} and \,\ref{eqa:sft} denote that only optimizer model $\bm{M_o}$ is included in preference optimization and supervised fine-tuning approaches. However, discrimination and instruction answering capacities are required in our iterative self-rewarding training, as shown in equations\,\ref{eqa:ipl_judge}. Hence, we simply re-use the collected data in the top of Figure\,\ref{fig:diversification}. 15k naive prompts $\bm{x^n}$ and ground truth responses $\bm{y^n}$ are paired as additional instruction following data of IPL, shown in the right middle part. Meanwhile, we set a new meta prompt for the inner discrimination of IPL, shown in the right corner of Figure\,\ref{fig:diversification}. Another 15k naive prompts $\bm{x^n}$, ground truth responses $\bm{y^n}$ and corresponding optimized prompts $\bm{x^{o-}}$ and $\bm{x^{o+}}$ are re-used for data construction.

In summary, we diversify the collected 30k raw data into eight various formats, for bridging the gaps between training and inference. Particularly, we add 15k instruction following data and 15k discrimination data for IPL, by re-using the raw data.

\section{HyperParameters and Training Cost}
\label{sec:app_cost}
We report hyperparameters and training cost in Table\,\ref{tab:cost}. We fine-tune all models on same computational node server. We only use single node for fine-tuning 13B models, but use 4 nodes for 70B models. While the batch size varies, we maintain a consistent global batch size on every node. In terms of different fine-tuning strategies, most hyperparameters are shared. The only difference lies on the learning rate, as we find large learning rate for DPO and IPO will cause collapse. What's more, all preference learning approaches relay on one external hyperparameter $\bm{\beta}$. We set its value as 0.01 empirically\,\cite{tunstall2023zephyr}. 

The neural optimizer that we adopted is \emph{AdamW}\,\cite{loshchilov2017decoupled} and the scheduler employed is \emph{WarmupDecayLR}\,\cite{goyal2017accurate}. We incorporate Deepspeed\,\cite{ren2021zero} and Flash Attention\,\cite{dao2022flashattention} to improve the training efficiency. It worth to mention that IPL approaches takes more than double training time per epoch compared with regular preference optimization approaches, since we increment additional instruction following data and discrimination data, as well as self-rewarding data updates.

\begin{table}[t]
\resizebox{\columnwidth}{!}{%
  \begin{tabular}{c|c|c}
    \toprule
    \quad & \textbf{Tulu2 13B} & \textbf{Tulu2 70B}\\
    \midrule
    \textbf{Nodes} & 1 & 4\\
    \midrule
    \textbf{Batch} & 8 & 2\\
    \midrule
    \textbf{Accumulations} & 2 & 8\\
    \midrule
    \textbf{HyperParams} & \multicolumn{2}{c}{\makecell{Epoch=3, Seq Len=2048, SFT Lr=2e-5, \\Else Lr=5e-7, Warmup Ratio=0.1, Beta=0.01, \\Gen TopP=0.95, Gen Temperature=0.8}}\\
    \midrule
    \textbf{Train (1 epoch / 3W)} & 2.5h & 2.5h\\
    \bottomrule
  \end{tabular}}
    \caption{Hyperparameters and training cost.}
  \label{tab:cost}
\end{table}

\begin{algorithm}[!t]
    \begin{algorithmic}[1]
        \STATE \textbf{Input require}: Total number of iterations $\bm{E}$,
        \STATE Optimizer model $\bm{M_o}$,
        \STATE \textbf{Initilization}: Naive Prompt $\bm{x^n}$,
            \STATE Naive Response $\bm{\hat{y}^n}$,
            \STATE Ground Truth Response $\bm{y}^n$,
        \FOR{$\bm{e}$ in $\bm{E}$}
            \IF{$\bm{e}$ > 1}
                \STATE New prompt $\bm{x^{n+}}=\bm{M_o}(\bm{x^n},\bm{\hat{y}^n},\bm{y}^n)$
                \IF{$\bm{M_o}(\bm{x^{n+}}, \bm{y^n}) \succ \bm{M_o}(\bm{x^n}, \bm{y^n})$}
                    \STATE New response $\bm{\hat{y}^{n+}}=\bm{M_o}(\bm{x^{n+}})$
                    \STATE Update $\bm{x^{n}}=\bm{x^{n+}}$
                    \STATE Update $\bm{\hat{y}^{n}}=\bm{\hat{y}^{n+}}$
                \ENDIF
            \ENDIF
            \STATE Update $\bm{M_o}$ with DPO or IPO loss.
        \ENDFOR
    \end{algorithmic}
    \caption{Self-rewarding IPL Algorithm.}
  \label{alg:ipl}
\end{algorithm}

\begin{table*}[!t]
\resizebox{\textwidth}{!}{%
  \begin{tabular}{c|c|cc|ccc|c}
    \toprule
    \quad & \quad & \multicolumn{2}{c|}{\textbf{Generation}} & \multicolumn{3}{c|}{\textbf{Multi-choice}} & \quad\\
    \toprule
    \quad & \textbf{FIPO Optimizer} & \textbf{GSM8K (3)} & \textbf{BBH (3)} & \textbf{PiQA (3)} & \textbf{CosmosQA (5)} & \textbf{MMLU (5)} & \textbf{Weighted Avg.}\\
    \midrule
    \quad & Naive & 24.77 & 36.21 & 73.35 & 51.17 & 51.22 & 47.79 \\
    \midrule
    SFT & 13B & 19.91 & 31.04 & 71.61 & 47.68 & 49.57 & 44.93 \\
    \quad & 70B & 21.43 & 32.92 & 74.39 & 49.97 & 51.55 & 46.96 \\
    \midrule
    DPO & 13B & 17.17 & 30.90 & 68.50 & 44.87 & 46.64 & 42.69 \\
    \quad & 70B & \textbf{27.74} & 35.56 & 74.17 & 54.93 & 52.73 & 49.07 \\
    \midrule
    \quad & 70B-e1 & 25.38 & 33.86 & 74.39 & 52.42 & 52.44 & 48.12 \\
    IPL-DPO & 70B-e2 & 24.28 & 36.10 & 73.84 & 51.22 & 52.04 & 48.23 \\
    \quad & 70B-e3 & 25.13 & 35.25 & 74.95 & 50.46 & 52.12 & 48.10 \\
    \midrule
    IPO & 13B & 18.42 & 33.55 & 72.03 & 49.14 & 48.20 & 44.93 \\
    \quad & 70B & 25.00 & 39.21 & 76.84 & 56.01 & 54.29 & 50.94 \\
    \midrule
    \quad & 70B-e1 & 25.99 & 34.09 & 74.66 & 53.62 & 52.33 & 48.30 \\
    IPL-IPO & 70B-e2 & 26.70 & 38.07 & 76.28 & 54.38 & 54.81 & 50.81 \\
    \quad & 70B-e3 & 26.67 & 39.60 & \textbf{77.11} & \textbf{56.71} & \textbf{56.02} & \textbf{52.13} \\
    \midrule
    IPL-IPO-gen & 70B-e3 & 22.72 & \textbf{41.91} & 74.53 & 53.60 & 52.74 & 50.23\\
    IPL-IPO-partial & 70B-e3 & 23.08 & 40.03 & 76.22 & 54.29 & 51.99 & 49.59\\
    \bottomrule
  \end{tabular}}
    \caption{Complete evaluation results of different fine-tuning strategies, tested by the same Tulu2-7B model. The ``\emph{-eN}'' notation refers to the ``\emph{N}''-th round of iteration in IPL training (e.g., 13B-e2 refers to the second iteration of fine-tuning the 13B model). ``\emph{-gen}'' and ``\emph{-partial}'' refer to two ablation experiments in accordance with Section\,\ref{sec:fine_analyasis}.}
  \label{tab:all_finetune}
\end{table*}

\section{Additional details of IPL}
\label{app:ipl}
We first report the algorithmic narrative of IPL in Algorithm\,\ref{alg:ipl}. In consistent with Equations\,\ref{eqa:ipl}, \,\ref{eqa:ipl_op}, and \,\ref{eqa:ipl_judge}, we first generate a new prompt $\bm{x^{n+}}$ with the optimizer model $\bm{M_o}$, then ask the optimizer also to judge if the newly optimized prompt is superior than the naive prompt $\bm{x^n}$. We takes the better prompt as the final one, and then generate a new task response $\bm{\hat{y}^{n+}}$ using the optimizer itself for next iteration. It should be noted that we start such self-rewarding updates after one epoch of warm-up.

we report the accuracy of fine-tuned discrimination ability and proportions of more rigours samples involved in IPL in Table\,\ref{tab:add_ipl}. We can notice that the optimizer can easily handle the binary discrimination task with simultaneous training on the additional discrimination data. We use 5\% training data as the validation set, and find 100\% classification accuracy. Based on this, the optimizer accurately update naive prompts $\bm{x}^n$ with self-generate new prompts $\bm{x}^{n+}$, introduced from line 9 to line 11 in Algorithm\,\ref{alg:ipl}. Nevertheless, the optimizer model gradually update new superior prompt samples for dynamic optimization with a conservative attitude, as shown in the ``\emph{Selection}'' column. We observe only 2.4\% naive prompts and responses samples are upgraded in the end, while the self-game approach significantly promotes the optimization of downstream test prompts, as mentioned in Table\,\ref{tab:all_finetune}.

\begin{table}[!t]
\resizebox{\columnwidth}{!}{%
  \begin{tabular}{l|c|c|c}
    \toprule
    \quad & \textbf{Weighted Avg.} & \textbf{Dis. Accuracy} & \textbf{Selection} \\
    \midrule
    Naive & 47.80 & N/A & N/A \\
    \midrule
    IPL-IPO-70B-e1 & 48.30 & N/A & N/A \\
    \midrule
    IPL-IPO-70B-e2 & 50.81 & 100\% & 1.25\% \\
    \midrule
    IPL-IPO-70B-e3 & 52.13 & 100\% & 2.40\% \\
    \bottomrule
  \end{tabular}}
    \caption{Analysis regards to details of IPL. The ``\emph{-eN}'' notation refers to the ``\emph{N}''-th round of iteration in IPL training. The selection refers to the data proportion that be updated with newly generated prompts. The first epoch is for warm-up, therefore there are not scores.}
  \label{tab:add_ipl}
\end{table}

\label{app:bbh}
\begin{table*}[!t]
\resizebox{\textwidth}{!}{%
  \begin{tabular}{c|c|c|c|c|c|c|c|c}
    \toprule
    \quad & \quad & \multicolumn{6}{c|}{\textbf{6 BBH Tasks following PromptAgent Work\,\cite{wang2023promptagent}}} & \quad \\
    \toprule
    \textbf{Downstream LLM} & \textbf{Optimizer} & \textbf{Penguins} & \textbf{Geometry} & \textbf{Epistemic} & \textbf{Object Counting} & \textbf{Temporal} & \textbf{Causal Judgment} & \textbf{Average}\\
    \toprule
    \quad & Naive & 0.436 & 0.536 & 0.611 & 0.318 & 0.362 & 0.547 & 0.468\\
    \quad & APE & 0.402 & 0.499 & 0.680 & 0.347 & 0.440 & 0.583 & 0.492\\
    Llama2-7B & PromptAgent & 0.426 & 0.544 & \textbf{0.775} & 0.312 & 0.643 & 0.547 & 0.541\\
    \,\cite{touvron2023llama} & GPT-4 & 0.444 & \textbf{0.578} & 0.640 & \textbf{0.392} & 0.536 & 0.598 & 0.531\\
    \quad & FIPO & \textbf{0.458} & 0.542 & 0.754 & 0.350 & \textbf{0.683} & \textbf{0.616} & \textbf{0.567}\\
    \midrule
    \quad & Naive & 0.336 & 0.089 & 0.581 & \textbf{0.690} & 0.155 & 0.832 & 0.447\\
    \quad & APE & 0.313 & 0.400 & 0.627 & 0.525 & 0.189 & 0.882 & 0.489\\
    Tulu2-7B & PromptAgent & 0.470 & 0.350 & \textbf{0.691} & 0.577 & 0.176 & 0.882 & 0.524\\
    \,\cite{ivison2023camels} & GPT-4 & \textbf{0.542} & 0.397 & 0.603 & 0.674 & 0.176 & 0.888 & 0.547\\
    \quad & FIPO & 0.525 & \textbf{0.415} & 0.644 & 0.637 & \textbf{0.210} & \textbf{0.891} & \textbf{0.554}\\
    \midrule
    \quad & Naive & 0.752 & 0.460 & 0.720 & 0.499 & 0.912 & 0.651 & 0.666\\
    \quad & APE & 0.740 & \textbf{0.533} & 0.713 & 0.530 & 0.897 & 0.643 & 0.676\\
    Llama3-70B-Instruct & PromptAgent & 0.748 & 0.501 & 0.667 & 0.524 & \textbf{0.946} & 0.687 & 0.687\\
    \,\cite{dubey2024llama} & GPT-4 & 0.756 & 0.478 & \textbf{0.726} & 0.472 & 0.912 & 0.644 & 0.665\\
    \quad & FIPO & \textbf{0.765} & 0.503 & 0.678 & \textbf{0.539} & 0.923 & \textbf{0.716} & \textbf{0.692}\\
    \midrule
    \quad & Naive & 0.593 & \textbf{0.540} & 0.698 & 0.602 & 0.879 & 0.767 & 0.680\\
    \quad & APE & 0.614 & 0.496 & 0.718 & 0.617 & 0.888 & \textbf{0.790} & 0.687\\
    Qwen2-72B-Instruct & PromptAgent & \textbf{0.658} & 0.533 & 0.717 & 0.622 & \textbf{0.898} & 0.779 & 0.722\\
    \,\cite{yang2024qwen2} & GPT-4 & 0.630 & \textbf{0.540} & 0.700 & \textbf{0.642} & 0.876 & 0.759 & 0.691\\
    \quad & FIPO & 0.617 & 0.522 & \textbf{0.792} & 0.591 & 0.836 & 0.768 & \textbf{0.731}\\
    \midrule
    \quad & Naive & 0.595 & 0.227 & 0.452 & 0.612 & 0.720 & 0.470 & 0.513\\
    \quad & APE & 0.747 & 0.490 & 0.708 & 0.716 & 0.856 & 0.570 & 0.681\\
    GPT-3.5-turbo & PromptAgent & \textbf{0.797} & 0.670 & 0.806 & \textbf{0.860} & \textbf{0.934} & 0.670 & \textbf{0.790}\\
    (In-box Tester) & GPT-4 & 0.632 & 0.589 & 0.840 & 0.647 & 0.821 & 0.550 & 0.680\\
    \quad & FIPO & 0.614 & \textbf{0.738} & \textbf{0.865} & 0.660 & 0.756 & \textbf{0.759} & 0.732\\
    \midrule
    \quad & Naive & 0.859 & 0.472 & 0.820 & 0.680 & \textbf{0.990} & 0.750 & 0.762\\
    \quad & APE & 0.862 & 0.691 & 0.844 & 0.650 & 0.988 & 0.749 & 0.797\\
    GPT-4o & PromptAgent & 0.855 & 0.750 & 0.886 & \textbf{0.733} & 0.982 & 0.712 & 0.820\\
    \quad & GPT-4 & 0.840 & 0.737 & \textbf{0.925} & 0.692 & 0.982 & 0.700 & 0.813\\
    \quad & FIPO & \textbf{0.876} & \textbf{0.810} & 0.891 & 0.684 & 0.965 & \textbf{0.840} & \textbf{0.844}\\
    \bottomrule
  \end{tabular}}
    \caption{Comparison between FIPO optimizer, previous methods and GPT-4's prompt optimization.}
  \label{tab:sota}
\end{table*}

\section{Entire fine-tuning results}
\label{sec:app_finetune}
We present entire fine-tuning results of our various experiments in Table\,\ref{tab:all_finetune}. Apart from conclusions reported in Section\,\ref{sec:fine_analyasis}, we further summarize several findings: (1) When fine-tuned with SFT approaches, 70B models perform better than 13B models as expected. However, all SFT models perform worse than the naive baseline; (2) In terms of two basic preference optimization training (DPO and IPO), we observe that 70B models are still superior to 13B models, and only the former ones obtains super-human results; (3) When it comes to our IPL approaches, there are differences when combined with DPO or IPO accordingly. IPL-DPO keeps a similar conclusion with DPO. As for IPL-IPO, we particularly find its full training works the best, and obtain an obvious growing trend, exhibiting a steeper upward growth of 2\% per epoch. We suppose the regularized design in Equation\,\ref{eqa:ipo} contributes to stable improvements.

\section{Detailed results of 6 BBH subsets}
We provide the performance of various optimizers on different downstream testers, including the evaluation results of 6 BBH tasks in Table\,\ref{tab:sota}. Here are some key points of the result analysis:

\textbf{Overall Performance}. The FIPO optimizer performs well in most combinations, usually leading in each task and average score. PromptAgent and GPT-4 optimizers perform well on some specific tasks, but are slightly inferior to FIPO overall.

\textbf{Optimizer Comparison}. When comparing different optimizers, FIPO outperforms other methods across multiple downstream LLMs. While PromptAgent occasionally achieves high scores in individual tasks, its average performance is generally lower than FIPO. Similarly, the GPT-4 optimizer shows competitive results in certain scenarios but does not maintain the same level of consistency as FIPO. On the other hand, the Naive prompt and APE optimizer often lag behind in most tasks. For example, on GPT-4o, APE's average score is 0.797, lower than FIPO's 0.844.

\textbf{Task-specific Performance}. FIPO excels in many tasks. In the "Temporal" and "Causal Judgment" tasks under Llama2-7B, it achieved the highest scores of 0.683 and 0.616, respectively. It also secured the top scores of 0.525 and 0.891 in the "Penguins" and "Causal Judgment" tasks under Tulu2-7B. For GPT-4o, FIPO achieved the highest scores in the "Geometry" and "Causal Judgment" tasks, with 0.810 and 0.840, respectively. Similarly, in the "Epistemic" task under Qwen2-72B-Instruct, FIPO led with a score of 0.792. 

In summary, the FIPO optimizer performed best on a variety of downstream LLMs, especially on GPT-4o and Llama3-70B-Instruct, where its stability and efficiency were remarkable. The PromptAgent and GPT-4 optimizers also performed well on specific tasks, but were not as comprehensive as FIPO overall. The Naive prompt and APE optimizer performed relatively poorly.

\section{Specific ablation discussion on BBH}
\label{app:ablation}
We notice that the BBH dataset behaves significantly different from other datasets in the ablation experiments, as shown by the last lines of Table\,\ref{tab:all_finetune}. The corrupted optimizer IPL-IPO-gen and IPL-IPO-partial gain suboptimal performances across all benchmarks, except BBH. We suppose this may be because the BBH contains a large number of symbolic reasoning tasks, and this unique pattern is not found in other testing datasets.

For instance, one of the typical subtask in BBH is ``Name a Geometry'', which asks to ``name geometric shapes from their SVG paths''. A typical example is listed here: ``\emph{This SVG path element <path d="M 59.43,52.76 L 75.49,27.45 L 54.92,4.40 M 54.92,4.40 L 23.70,7.77 L 15.15,42.15 L 34.51,57.44 L 59.43,52.76"/> draws a }''. Therefore, we suspect that when further using discrete natural language instructions to diversify our training data, this may affect subtasks in BBH that involve specific symbolic reasoning patterns.

Nevertheless, when examining the six representative subtasks in BBH, FIPO still achieves a clear advantage over other leading prompt optimization methods, as shown by the Table\,\ref{tab:sota}.

\section{Cost comparison: FIPO vs. others}
\label{app:cost}
In terms of overall cost against other aforementioned optimization approaches, shown in Table\,\ref{tab:cost_diff}, building FIPO optimizer will have \$300 one-time collecting cost and 10 hours training cost, which equals to 75 usages of GPT-4. However, once FIPO models are prepared, we save a lot of time as the optimization can be done locally. FIPO leads to less total costs when optimizing over massive prompts.

In specific, we explain the cost number in Table\,\ref{tab:cost_diff} column by column. The first column refers to the cost of constructing data. Since APE, PromptAgent and directly using GPT-4 are all training-free methods, they do not have the cost of constructing training data. Similarly, the second column refers to the time cost of training, and only our local fine-tuned FIPO optimizer covers that. As for the inference cost in the third and fourth columns, we report the inference cost of the entire representative ``\emph{penguins\_in\_a\_table}'' subtask in BBH, which includes 149 test samples. In particular, it worth to note that the inference cost of PromptAgent is consistent with its official report\footnote{\url{https://github.com/XinyuanWangCS/PromptAgent/}}. In addition, as the only local optimizer, FIPO also includes a one-time inference deployment cost. However, in our practice, since the time for renting a server and deploying a model does not exceed 3 minutes in total, this part of the cost is basically negligible.

\section{Overwhelmed Cheating Notes}
\label{sec:app_cheat}
\begin{figure}[!t]
  \centering
  \includegraphics[width=1.0\linewidth]{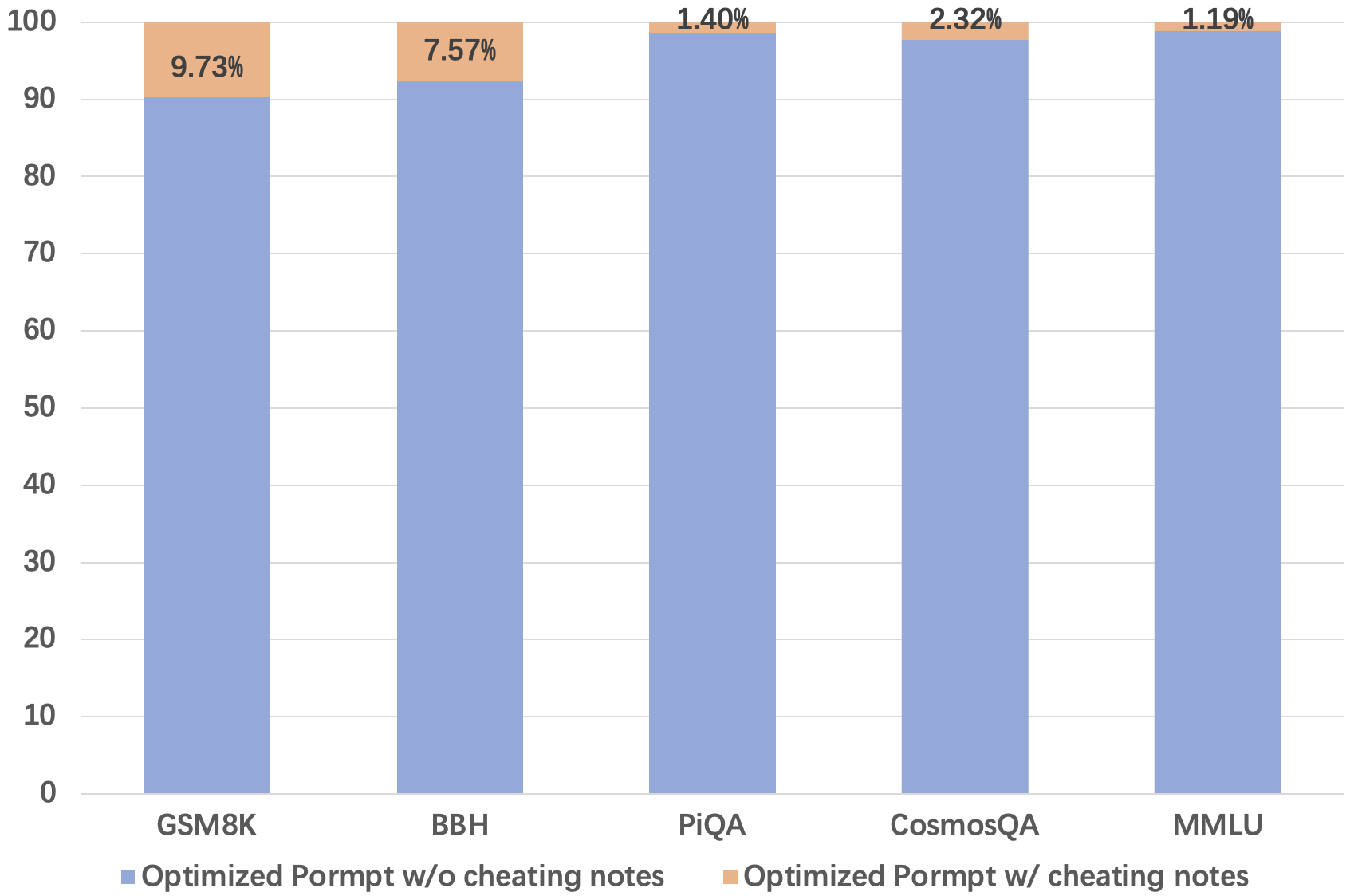}
  \caption{Analysis of overwhelmed cheating notes on 256 random optimized prompts for each benchmark.}
  \label{fig:cheating}
\end{figure}

We provide a proportion visualization of analyzing the possible over-optimized prompts in each testing benchmark, manually checking with 256 random samples, in Figure\,\ref{fig:cheating}. The results show that mathematical questions in GSM8K and BBH are more likely to be added cheating notes of final answers, while other types of testing data receive moderate prompt optimization. Nevertheless, even taking into account this factor, the conclusion that our local optimizer improves the overall performance of general downstream generators by optimizing prompts reasonably does not change (\S\,\ref{sec:sota}).

\begin{table}[t]
\resizebox{\columnwidth}{!}{%
  \begin{tabular}{l|c|c|c|c}
    \toprule
    \quad & \quad & \quad & \multicolumn{2}{|c}{Inference Per Test} \\
    \quad & \textbf{Dataset} & \textbf{Training} & \textbf{Fee} & \textbf{Time} \\
    \midrule
    \textbf{APE} & \$0 & \$0 & \$5 & 2h \\
    \midrule
    \textbf{PromptAgent} & \$0 & \$0 & \$5 & 2h \\
    \midrule
    \textbf{GPT-4} & \$0 & \$0 & \$4 & 1h \\
    \midrule
    \textbf{FIPO (IPL-IPO-70B)} & \$300 & \$60 & \$0 & 30s \\
    \bottomrule
  \end{tabular}}
    \caption{Cost report of different methods.}
  \label{tab:cost_diff}
\end{table}

\section{Employed Meta Prompts}
\label{sec:app_meta}

We report all employed meta prompts in this section as reference for potential future researches. Please refer to Table\,\ref{tab:collection}, \,\ref{tab:divers} and \,\ref{tab:discrim} for data collection meta prompt, data diversification meta prompt, and discrimination meta prompt, respectively.

\begin{table*}[!t]
\resizebox{\textwidth}{!}{%
  \begin{tabular}{p{\textwidth}}
    \toprule
    \textbf{Data collection meta prompt}\\
    \midrule
    You are an expert of prompt optimization.\\
    \quad \\
    \`{}\`{}\`{} \\
    Sliver Prompt:\\
    SP\\
    \`{}\`{}\`{} \\
    \quad \\
    \`{}\`{}\`{} \\
    Sliver Response:\\
    SR\\
    \`{}\`{}\`{} \\
    \quad \\
    \`{}\`{}\`{} \\
    Golden Response:\\
    GR\\
    \`{}\`{}\`{} \\
    \quad \\
    \`{}\`{}\`{} \\
    Task Introduction:\\
    Based on the Silver Prompt, optional Silver Response and optional Golden Response, perform the following actions:\\
    1 – The optional Sliver Response was generated by an AI based on the Silver Prompt. Please help modify the Silver Prompt to Golden Prompt that can obtain a more correct response, in reference to the optional Golden Response.\\
    2 - When building the Golden Prompt, you can consider several aspects, such as: (1) A roleplay leading sentence to adapt the AI to the task-specific scenario; (2) Details of task characteristics, for instance, the task could be a question answering task, a dialogue task, or a summarization task, etc; (3) Further clarification of the task information, especially some ambiguous terms; (4) A more detailed solution guidance, such as step-by-step plans, handlings of exceptions, special priorities or constraints, etc; (5) Any specific requirements for the response, such as the length, the format, the style, the tone, the language, etc.\\
    3 - Show me only the Golden Prompt, do not contain any other content.\\
    \`{}\`{}\`{} \\
    \quad \\
    Golden Prompt:\\
    \bottomrule
  \end{tabular}}
    \caption{The meta prompt used to harness data from GPT-3.5-turbo and GPT-4 APIs. During harnessing, we replace placeholders ``\emph{SP}'', ``\emph{SR}'' and ``\emph{GR}'' with actual naive prompt $\bm{x^n}$, naive response $\bm{\hat{y}^n}$ and ground truth response $\bm{y}^n$ from each seed data sample, respectively.}
  \label{tab:collection}
\end{table*}

\begin{table*}[!t]
\resizebox{\textwidth}{!}{%
  \begin{tabular}{p{\textwidth}}
    \toprule
    \textbf{Data diversification meta prompt}\\
    \midrule
    You are an expert of prompt optimization.\\
    \quad \\
    \`{}\`{}\`{} \\
    Sliver Prompt:\\
    SP\\
    \`{}\`{}\`{} \\
    <Optional Responses>\\
    \quad \\
    The optional Sliver Response was generated by an AI based on the Silver Prompt. Please help modify the Silver Prompt to Golden Prompt that can obtain a more correct response, in reference to the optional Golden Response. The Golden Prompt should be instructive, concise and strictly faithful to any factual information in the Silver Prompt. The length of the Golden Prompt should be less than GN words. Only give me the content of Golden Prompt, do not contain any other information (e.g., your response of the Golden Prompt, any postfix like ``Golden Prompt'', etc.).\\
    \midrule
    \textbf{Flexible meta prompt for optional naive response}\\
    \midrule
    \quad \\
    \`{}\`{}\`{} \\
    Sliver Response:\\
    SR\\
    \`{}\`{}\`{} \\
    \midrule
    \textbf{Flexible meta prompt for optional ground truth response}\\
    \midrule
    \quad \\
    \`{}\`{}\`{} \\
    Golden Response:\\
    GR\\
    \`{}\`{}\`{} \\
    \bottomrule
  \end{tabular}}
    \caption{The meta prompt used for data diversification. During diversification, we replace placeholders ``\emph{SP}'', ``\emph{SR}'' and ``\emph{GR}'' with actual naive prompt $\bm{x^n}$, naive response $\bm{\hat{y}^n}$ and ground truth response $\bm{y}^n$, respectively. Meanwhile, flexible prompts of optional naive response and ground truth response are inserted by probability as introduced in Figure\,\ref{fig:finetuning} and \,\ref{fig:diversification}. Moreover, we add length suggestion with ``\emph{GN}'' signal, using the length of chosenl optimized prompt label $\bm{x^{o+}}$.}
  \label{tab:divers}
\end{table*}

\begin{table*}[!t]
\resizebox{\textwidth}{!}{%
  \begin{tabular}{p{\textwidth}}
    \toprule
    \textbf{Discrimination meta prompt}\\
    \midrule
    You are an expert of prompt discrimination.\\
    \quad \\
    \`{}\`{}\`{} \\
    Raw Prompt:\\
    RP\\
    \`{}\`{}\`{} \\
    \quad \\
    \`{}\`{}\`{} \\
    New Prompt A:\\
    PA\\
    \`{}\`{}\`{} \\
    \quad \\
    \`{}\`{}\`{} \\
    New Prompt B:\\
    PB\\
    \`{}\`{}\`{} \\
    \quad \\
    \`{}\`{}\`{} \\
    Golden Response:\\
    GR\\
    \`{}\`{}\`{} \\
    \quad \\
    New Prompt A and New Prompt B are optimized from Raw Prompt. Please judge which prompt is more loyal to the factual information of Raw Prompt, and is more desirable for an AI to generate the Golden Response. Only answer with A or B.\\
    \bottomrule
  \end{tabular}}
    \caption{The meta prompt used for creating discrimination data and application of discrimination during IPL training. The creation of discrimination data has been introduced in the right part of Figure\,\ref{fig:diversification}. In terms of inner discrimination of IPL, we replace placeholders ``\emph{RP}'', ``\emph{PA}'', ``\emph{PB}'' and ``\emph{GR}'' with naive prompt $\bm{x^n}$, naive prompt $\bm{x^n}$, newly optimized prompt $\bm{x^{n+}}$ and ground truth response $\bm{y}^n$, respectively.}
  \label{tab:discrim}
\end{table*}

\end{document}